\documentclass{sig-alternate-05-2015}
\usepackage{url}
\usepackage[notheorems]{nkahmed-math}

\usepackage{amsmath}
\usepackage{amsfonts}
\usepackage{amssymb}

\usepackage{verbatim}
\usepackage{wrapfig}
\usepackage{booktabs}
\usepackage{graphicx}
\usepackage{multirow}
\usepackage{rotate}
\usepackage{subfigure}
\usepackage{epstopdf}
\usepackage{paralist}
\usepackage{url}
\usepackage{graphicx}
\usepackage{wrapfig}
\usepackage{xspace}
\usepackage{xcolor}
\usepackage{epsfig}

\usepackage{graphicx}

\newcommand{\todo}[1]{}

\newcommand{\incomplete}[1]{}

\usepackage{tabularx}
\usepackage{booktabs}

\usepackage{relsize}
\usepackage{algorithm}
\usepackage{algorithmicx}
\usepackage{algpseudocode}

\algblockdefx[parallel]{ParFor}{EndPar}[1][]{$\textbf{parallel for}$ #1 $\textbf{do}$}{$\textbf{end parallel}$}
\algrenewcommand{\alglinenumber}[1]{\fontsize{6.5}{7}\selectfont#1}
\algtext*{EndPar}

\algblockdefx[parallel]{parfor}{endpar}[1][]{$\textbf{parallel for}$ #1 $\textbf{do}$}{$\textbf{end parallel}$}
\algrenewcommand{\alglinenumber}[1]{\scriptsize#1:}

\usepackage{nicefrac}

\algnotext{EndFor}
\algnotext{EndIf}
\algnotext{EndProcedure}

\usepackage{array}
\newcolumntype{P}[1]{>{\centering\arraybackslash}p{#1}}
\newcolumntype{M}[1]{>{\centering\arraybackslash}m{#1}}

\usepackage{listings}

\newcommand{\eol}{\end{enumerate}\setlength{\itemsep}{-\parsep}}

\relax

\usepackage{wrapfig}
\usepackage{xspace}

\usepackage{csquotes}

\newlength{\commentWidth}
\newcommand{\bspacing}{\begin{spacing}{1.4}}
\newcommand{\espacing}{\end{spacing}}

\definecolor{plotblue}{RGB}	{30,144,255}
\definecolor{plotgreen}{RGB}	{50,205,50}
\definecolor{plotred}{RGB}	{220,20,60}

\definecolor{myyellow}{RGB}{255,255,204}
\definecolor{myred}{RGB}{255,204,204}
\definecolor{myblue}{RGB}{204, 255, 255}
\definecolor{mygreen}{RGB}{204, 255, 204}

\definecolor{gray}{RGB}{150,150,150}
\definecolor{theblue}{RGB}{0,0,180}

\usepackage{textcase}
\usepackage{soul}

\newcommand*\hrulefillvar[1][0.4pt]{\leavevmode\leaders\hrule height#1\hfill\kern0pt}

\newcommand{\be}{\begin{equation}}
\newcommand{\ee}{\end{equation}}
\newcommand{\bea}{\begin{eqnarray}}
\newcommand{\eea}{\end{eqnarray}}
\newcommand{\bit}{\begin{itemize}}
\newcommand{\eit}{\end{itemize}}

\definecolor{lightgray}{rgb}{0.93,0.93,0.93}

\definecolor{lightblue}{rgb}{0.5,0.90,1.0}
\definecolor{lightgreen}{rgb}{0.5,0.92,0.5}
\definecolor{lightred}{rgb}{0.98,0.5,0.5}
\definecolor{lightyellow}{rgb}{1,0.90,0.40}

\usepackage{morefloats}
\usepackage{booktabs}
\usepackage{comment}

\newcommand{\eat}[1]{}

\definecolor{myyellow}{RGB}{255,255,204}
\definecolor{myred}{RGB}{255,204,204}
\definecolor{myblue}{RGB}{0,200,255}
\definecolor{mygreen}{RGB}{80,220,80}

\newcommand{\eg}{\emph{e.g.}}
\newcommand{\ie}{\emph{i.e.}}

\newcommand{\wrt}{\emph{w.r.t.}\ }

\newcommand{\abs}[1]{\left|#1\right|}

\newcommand{\inner}[2]{\langle #1,#2 \rangle}

\algblockdefx[parallel]{parallelfor}{parallelend}
[1][]{\textbf{parallel for} #1}
{\textbf{end parallel}}

\RequirePackage{mathtools}
\RequirePackage{color}
\RequirePackage{xcolor}
\RequirePackage{listings}

\newcommand{\ds}\displaystyle
\newcommand{\mbb}\mathbb
\newcommand{\mc}\mathcal
\newcommand{\del}\nabla

\newcommand{\beqstar}{\begin{eqnarray*}}
\newcommand{\eeqstar}{\end{eqnarray*}}

\definecolor{thegreen}{rgb}{0,.5,0}
\definecolor{idea}{rgb}{0,.6,0.1}
\definecolor{problem}{rgb}{0.7,0,0.1}
\definecolor{comment-green}{rgb}{0,.3,0}
\definecolor{theblue}{rgb}{0,0,.8}
\definecolor{light-gray}{gray}{0.98}
\definecolor{comment-color}{rgb}{0,0,.8}
\definecolor{string-color}{rgb}{0,.75,0}
\definecolor{border-blue}{rgb}{0,0,.6}

\usepackage{epsfig}
\usepackage[T1]{fontenc}
\usepackage{ragged2e}
\newcolumntype{H}{>{\setbox0=\hbox\bgroup}c<{\egroup}@{}}

\usepackage{pifont}

\definecolor{orange}{rgb}{1,0.5,0}
\definecolor{gray}{RGB}{20,20,20}
\definecolor{greencm}{RGB}{0,153,0}

\usepackage{upgreek}

\usepackage{ifpdf}
\usepackage{xcolor}
\usepackage{paralist}
\usepackage{tabularx}
\usepackage{booktabs}
\usepackage{multirow}
\usepackage{graphicx}
\usepackage{sidecap}
\usepackage{mathdots}
\usepackage{lscape}
\makeatletter
\def\vcdots{\vbox{\baselineskip4\p@ \lineskiplimit\z@
    \kern3\p@\hbox{.}\hbox{.}\hbox{.}\kern3\p@}}
\makeatother
\usepackage{subfigure}
\usepackage{array}
\usepackage{url}
\usepackage{float}

\providecommand{\RR}{\mathbb{R}} 

\providecommand{\Std}{\mathbb{S}} 

\renewcommand{\argmax}{\operatornamewithlimits{\arg \; \max}}

\providecommand{\T}{\ensuremath{{T}}} 

\providecommand{\E}{\ensuremath{E}}

\providecommand{\e}{\ensuremath{e}}  		

\providecommand{\X}{\ensuremath{X}}  
\providecommand{\p}{\ensuremath{p}} 
\providecommand{\x}{\ensuremath{\sx}}  

\providecommand{\dmax}{\ensuremath{\Delta}} 

\providecommand{\p}{\ensuremath{p}}

\providecommand{\S}{\ensuremath{{S}}} 

\providecommand{\bigO}[1]{\ensuremath{\mathcal{O}(#1)}}

\providecommand{\P}{\ensuremath{\mathrm{P}}} 

\providecommand{\V}{\ensuremath{{\mV}}}
\providecommand{\U}{\ensuremath{{\mU}}}

\providecommand{\m}{\ensuremath{m}}
\providecommand{\n}{\ensuremath{n}}

\providecommand{\f}{\ensuremath{f}} 
\renewcommand{\r}{\ensuremath{r}} 

\providecommand{\D}[2]{\ensuremath{\mathbb{D}_{\phi}(#1 \| #2)}}

\renewcommand{\subsubsection}[1]{\medskip\noindent\textbf{{{#1}:\,}}}

\begin{document}
\title{Revisiting Role Discovery in Networks: \\From Node to Edge Roles}

\numberofauthors{4} 
\author{
\alignauthor
Nesreen K. Ahmed\\
       \affaddr{Intel Labs}\\
       \email{nesreen.k.ahmed@intel.com}
\alignauthor
Ryan A. Rossi\\
       \affaddr{Palo Alto Research Center}\\
       \email{rrossi@parc.com}
\alignauthor
Theodore L. Willke\\
\affaddr{Intel Labs}\\
\email{ted.willke@intel.com}
\and
Rong Zhou\\
       \affaddr{Palo Alto Research Center}\\
       \email{rzhou@parc.com}
}

\maketitle
\begin{abstract}
Previous work in network analysis has focused on modeling the mixed-memberships of node roles in the graph, but not the roles of edges. 
We introduce the \emph{edge role discovery problem} and present a generalizable framework for learning and extracting edge roles from arbitrary graphs automatically. 
Furthermore, while existing node-centric role models have mainly focused on simple degree and egonet features, this work also explores graphlet features for role discovery.
In addition, we also develop an approach for automatically learning and extracting important and useful edge features from an arbitrary graph.
The experimental results demonstrate the utility of edge roles for network analysis tasks on a variety of graphs from various problem domains.
\end{abstract}

\section{Introduction} \label{sec-intro}
\noindent
In the traditional graph-based sense, roles represent node-level connectivity patterns such as star-center, star-edge nodes, near-cliques or nodes that act as bridges to different regions of the graph. Intuitively, two nodes belong to the same role if they are ``similar" in the sense of graph structure. 
Our proposed research will broaden the framework for defining, discovering and learning network roles, by drastically increasing the degree of usefulness of the information embedded within rich graphs.

Recently, role discovery has become increasingly important for a variety of application and problem domains~\cite{borgatti2013book,hollandkathryn1983stochastic,arabie1978constructing,anderson1992building,rossi2015-tkde,lorrain1971structural,white1983graph} including descriptive network modeling~\cite{rossi2013dbmm-wsdm}, classification~\cite{rolX}, anomaly detection~\cite{rossi2013dbmm-wsdm}, and exploratory analysis (See~\cite{rossi2015-tkde} for other applications). 
Despite the (wide variety of) practical applications and importance of role discovery, existing work has only focused on discovering node roles (\eg, see\cite{anderson1992building,batagelj2004generalized,doreian2005generalized,nowicki2001estimation}).
We posit that discovering the roles of edges may be fundamentally more important and able to capture, represent, and summarize the key behavioral roles in the network better than existing methods that have been limited to learning only the roles of nodes in the graph.
For instance, a person with malicious intent may appear normal by maintaining the vast majority of relationships and communications with individuals that play normal roles in society. In this situation, techniques that reveal the role semantics of nodes would have difficulty detecting such malicious behavior since most edges are normal. 
However, modeling the roles (functional semantics, intent) of individual edges (relationships, communications) in the rich graph would improve our ability to identify, detect, and predict this type of malicious activity since we are modeling it directly. 
Nevertheless, existing work also have many other limitations, which significantly reduces the practical utility of such methods in real-world networks.
One such example is that the existing work has been limited to mainly simple degree and egonet features~\cite{rolX,rossi2013dbmm-wsdm}, see~\cite{rossi2015-tkde} for other possibilities.
Instead, we leverage higher-order network motifs (induced subgraphs) of size $k \in \{3, 4, \ldots\}$ computed from~\cite{pgd,ahmed2016kais} and other graph parameters such as the largest clique in a node (or edge) neighborhood, triangle core number, as well as the neighborhood chromatic, among other efficient and highly discriminative graph features.

{
\bigskip
\noindent
The main contributions are as follows:
\begin{itemize}
\item \textbf{Edge role discovery:} This work introduces the problem of edge role discovery and proposes a computational framework for learning and modeling edge roles in both static and dynamic networks. 

\item \textbf{Higher-order latent space model:} 
Introduced a higher-order latent role model that leverages higher-order network features for learning and modeling node and edge roles.
We also introduced graphlet-based roles and proposed feature and role learning techniques.

\item \textbf{Efficient and scalable:} All proposed algorithms are parallelized. 
Moreover, the feature and role learning and inference algorithms are linear in the number of edges.
\end{itemize}
}

\section{Higher-order Edge Role Model} \label{sec-framework}
\noindent
This section introduces our higher-order edge role model and a generalizable framework for computing edge roles based on higher-order network features.

\subsection{Initial Higher-order Network Features} \label{sec:higher-order-network-features}
Existing role discovery methods use simple degree-based features~\cite{rolX}.
In this work, we use graphlet methods~\cite{fanmod,rage,pgd} for computing higher-order network features based on induced subgraph patterns (instead of simply edge and node patterns) for discovering better and more meaningful roles.
Following the idea of feature-based roles~\cite{rossi2015-tkde}, we systematically discover an edge-based feature representation. 
\cite{fanmod,rage}
As initial features, we used a recent parallel graphlet decomposition framework proposed in \cite{pgd} to compute a variety of edge-based graphlet features of size $k=\{3,4,\ldots\}$.
Using these initial features, more representative, explainable, and novel features can be discovered.
See the relational feature learning template given in~\cite{rossi2015-tkde}.
As an aside, graphlets offer a way to generalize many existing feature learning systems (including those that have been used for learning and extracting roles).
We can generalize the above by introducing a generic k-vertex graphlet operator that returns counts and other statistics for any $k$-vertex induced subgraph where $k>2$.

\begin{table}[t!]
\caption{Summary of Bregman Divergences and update rules}
\vspace{1mm}
\label{table:summary-bregman-div}
\setlength{\tabcolsep}{2.2pt} 
\centering 
\fontsize{8.5}{10}\selectfont
\def\arraystretch{1.2}
\begin{tabularx}{1.0\linewidth}{@{}c ccc l@{}} 
\toprule
& $\phi(y)$ &
$\nabla^{2}\phi(y)$ & 
$\D{x}{x^{\prime}}$ &
\textbf{Update} \\
\midrule

\def\arraystretch{1.5}
\textbf{Fro.} &
$y^2 / 2$ &
$1$ & 
$(x-x^{\prime})^2 / 2$ & 
$v_{jk} = \frac{\sum_{i=1}^{\m} x^{(k)}_{ij} u_{ik}}{\sum_{i=1}^{\m} u_{ik} u_{ik}}$ \\

\textbf{KL} &
$y \log y$ &
$1 / y$ & 
$x \log \frac{x}{x^{\prime}} - x + x^{\prime}$ & 
$v_{jk} = \frac{\sum_{i=1}^{\m} x^{(k)}_{ij} u_{ik} / x^{\prime}_{ij}} {\sum_{i=1}^{\m} u_{ik} u_{ik} / x^{\prime}_{ij}}$ \\

\textbf{IS} & 
$- \log y$ &
$1 / y^2$ & 
$\frac{x}{x^{\prime}} - \log \frac{x}{x^{\prime}}$ & 
$v_{jk} = \frac{\sum_{i=1}^{\m} x^{(k)}_{ij} u_{ik} / x^{\prime}_{ij}{^2}}{\sum_{i=1}^{\m} u_{ik} u_{ik} / x^{\prime}_{ij}{^2}}$ \\

\bottomrule
\end{tabularx}
\end{table}

\subsection{Edge Feature Representation Learning} \label{sec:edge-feature-learning-and-extraction}
Learning important and practical representations automatically is useful for many machine learning applications beyond role discovery such as anomaly detection, classification, and descriptive modeling/exploratory analysis. 
These methods greatly reduce the engineering effort while also revealing important latent features that lead to better predictive performance and power of generalization. 
This section introduces a generalizable, flexible, and extremely efficient \emph{edge feature learning and inference framework} capable of automatically learning a representative set of edge features automatically.
The proposed framework and algorithms that arise from it naturally support arbitrary graphs including undirected, directed, and/or bipartite networks.
More importantly, our approach also handles attributed graphs in a natural way, which typically consist of a graph $G$ and a set of arbitrary edge and/or node attributes. 
The attributes typically represent intrinsic edge and node information such as age, location, gender, political views, textual content of communication between individuals, among other possibilities.

For edge feature learning and extraction, we introduce the notion of an edge neighbor.
Intuitively, given an edge $e_i=(v,u) \in E$, let $e_j=(a,b)$ be an edge neighbor of $e_i$ iff $a=v$, $a=u$, $b=v$, or $b=u$. 
Informally, $e_j$ is a neighbor of $e_i$ if $e_j$ and $e_i$ share a vertex. 
This definition can easily be extended for incorporating further $h$-distant neighbors.

The relational operators used to search the space of possible neighbor features at the current and previous learned feature layers include relational operators such as mean, sum, product, $\min$, $\max$, variance, L1, L2, and more generally, any (parameterized) similarity function including positive semidefinite functions such as the Radial Basis Function (RBF) $K\inner{\vx_i}{\vx_j} = \mathrm{exp}(-\nicefrac{\|\vx_i - \vx_j\|^{2}}{2\sigma^2})$, 
polynomial similarity functions of the form $K(\vx_i,\vx_j) = (a \cdot \inner{\vx_i}{\vx_j} + c)^{d}$, 
sigmoid neural network kernel $K(\vx_i,\vx_j) = \tanh(a \cdot \inner{\vx_i}{\vz_j} + c)$, among others.
The flexibility and generalizability of the proposed approach is a key advantage and contribution, $\ie$, many components are naturally interchangeable, and thus our approach is not restricted to only the relational operators mentioned above, but can easily leverage other application or problem domain specific (relational) operators\footnote{See~\cite{feature-extraction} for more details.}. 

Features are searched from the selected feature subspaces and the candidate features that are actually computed are pruned to ensure the set learned is as small and representative as possible capturing novel and useful properties.
We define a feature graph $G_f$ where the nodes represent features learned thus far among all current feature layers as well as any initial features which were not immediately pruned due to being redundant or non-informative \wrt the objective function\footnote{In general, the objective function can be either unsupervised or supervised.} whereas the edges encode dependencies between the features.
Further, the edges are weighted by the computed similarity/correlation (or distance/disagreement) measure, thus as $W_{ij} \rightarrow 1$ then the two features $f_i$ and $f_j$ are considered to be extremely similar (and thus possibly redundant), whereas $W_{ij} \rightarrow 0$ implies that $f_i$ and $f_j$ are significantly different. 
In this work, we use log-binning disagreement, though have also used Pearson correlation, among others.

After constructing the weighted feature graph, 
our approach has two main steps:
pruning noisy edges between features and 
the removal of redundant features (\ie, nodes in $G_f$).
To remove these spurious relationships in the feature graph, we use a simple adaptive sparsification technique.
The technique uses a threshold $\gamma$ which is adapted automatically at each iteration in the search procedure.
Once the spurious edges have been removed entirely from the feature graph, 
we then prune entire features that are found to be redundant. 
In other words, we discard vertices (i.e., features) from the feature graph that offer no discriminatory power. 
This can be performed by partitioning the feature graph in some fashion (\eg, connected components).
Note that once a feature is added to the set of representative features, it cannot be pruned. 
However, all features are scored at each iteration, including the representative features, and redundant features are discarded. 
If a feature is found to closely resemble one of the representative features, we prune it, and keep only the representative feature, as it is more primitive (discovered in a previous iteration). 
Our approach searches the space of features until one of the following stopping criterion are met:
(i) the previous iteration was unfruitful in discovering any novel features, or if
(ii) the maximum number of iterations is exceeded which may be defined by the user.

It is worth mentioning that we could have used an arbitrary node feature learning approach such as the one described in~\cite{rossi2015-tkde}. 
For instance, given a node feature matrix $\mZ \in \RR^{n \times f}$ (from one such approach), we can easily derive edge features from it (which can then be used for learning edge roles) by using one or more operators over the edges as follows: 
given an edge $e_k = (v_i,v_j) \in E$ with end points $v_i$ and $v_j$, one can simply combine the feature values $z_i$ and $z_j$ of $v_i$ and $v_j$, respectively, in some way, \eg, $x_k = z_i+z_j$ where $x_k$ is the resulting edge feature value for $e_k$.

\subsection{Learning Latent Higher-order Edge Roles}
\label{sec:edge-role-learning-and-inference} 
Let $\mX = \big[ x_{ij}\big] \in \RR^{\m \times \f}$ be a matrix with $\m$ rows representing edges and $\f$ columns representing arbitrary features\footnote{For instance, the columns of $\mX$ represent arbitrary features such as graph topology features, non-relational features/attributes, and relational neighbor features, among other possibilities.}.
More formally, given $\mX \in \RR^{\m \times \f}$, the edge role discovery optimization problem is to find $\U \in \RR^{\m \times \r}$ and $\V \in \RR^{\f \times \r}$ where $\r \ll \min(\m,\f)$ such that the product of two lower rank matrices $\U$ and $\V^{\T}$ minimizes the divergence between $\mX$ and $\mX^{\prime}=\U\V^{\T}$.
Intuitively, $\U \in \RR^{\m \times \r}$ represents the latent \emph{role mixed-memberships} of the edges whereas $\V  \in \RR^{\f \times \r}$ represents the contributions of the features with respect to each of the roles.
Each row $\vu_{i}^{\T} \in \RR^{\r}$ of $\U$ can be interpreted as a low dimensional rank-$\r$ embedding of the $i^{th}$ \emph{edge} in $\mX$.
Alternatively, each row $\vv_{j}^{\T} \in \RR^{\r}$ of $\V$ represents a $\r$-dimensional role embedding of the $j^{th}$ feature in $\mX$ using the same low rank-$r$ dimensional space.
Also, $\vu_k \in \RR^{\m}$ is the $k^{th}$ column representing a ``latent feature" of $\U$ and similarly $\vv_k \in \RR^{\f}$ is the $k^{th}$ column of $\V$.

For the higher-order latent network model, we solve:
\begin{align} \label{eq:edge-role-model-bregman-generalization}
\argmin_{(\U,\V) \in \mathcal{C}} \; \Bigl\{ \D{\mX}{\, \U\V^{T}} + \mathcal{R}(U,V) \Bigr\}
\end{align}
\noindent
where $\D{\mX}{\U\V^{T}}$ is an arbitrary Bregman divergence~\cite{bregman1967relaxation} between $\mX$ and $\U\V^T$.
Furthermore, the optimization problem in $\eqref{eq:edge-role-model-bregman-generalization}$ imposes hard constraints $\mathcal{C}$ on $\U$ and $\V$ such as non-negativity constraints $\U,\V \geq 0$ and $\mathcal{R}(U,V)$ is a regularization penalty. 
In this work, we mainly focus on solving $\D{\mX}{\U\V^{T}}$ under non-negativity constraints: 
\begin{align} \label{eq:edge-role-model-bregman-generalization-nmf}
\argmin_{\U \geq 0, \V \geq 0} \; \Bigl\{ \D{\mX}{\, \U\V^{T}} + \mathcal{R}(U,V) \Bigr\}
\end{align}
Given the edge feature matrix $\mX \in \RR^{\m \times \f}$, the edge role discovery problem is to find $\U \in \RR^{\m \times \r}$ and $\V \in \RR^{\f \times \r}$ such that 
\begin{align} \label{eq:approximation}
\mX \approx \mX^{\prime} = \U\V^{\T}
\end{align}
\noindent
To measure the quality of our edge mixed membership model, we use Bregman divergences:
\begin{align} \label{eq:approximation-quality} \nonumber
\sum_{ij} \D{x_{ij}}{x_{ij}^{\prime}}
= \sum_{ij} \big(\phi(x_{ij}) - \phi(x_{ij}^{\prime})   -  \ell(x_{ij}, x_{ij}^{\prime}) \big)
\end{align}
\noindent
where $\phi$ is a univariate smooth convex function and
\[
\ell(x_{ij}, x_{ij}^{\prime}) = \nabla\phi(x_{ij}^{\prime})(x_{ij} - x_{ij}^{\prime}), 
\]
\noindent
where $\nabla^{p}\phi(x)$ is the p-order derivative operator of $\phi$ at $x$. 
Furthermore, let $\mX - \mU\mV^{T} = \mX^{(k)} - \vu_k \vv_k^{T}$ denote the residual term in the approximation \eqref{eq:approximation} where $\mX^{(k)}$ is the k-residual matrix defined as:
\begin{align}\label{eq:k-residual-matrix}
\mX^{(k)} &= \mX - \sum\limits_{h\not=k} \vu_{h} \vv_{h}^{T} \\
&= \mX - \mU\mV^{T} + \vu_k \vv_k^{T}, \quad \text{for } k=1,\dots,\r
\end{align} 

We use a fast \emph{scalar block coordinate descent approach} that easily generalizes for heterogeneous networks~\cite{pcmf-snam16}.
The approach considers a single element in $\mU$ and $\mV$ as a block in the block coordinate descent framework.
Replacing $\phi(y)$ with the corresponding expression from Table~\ref{table:summary-bregman-div} gives rise to a fast algorithm for each Bregman divergence.
Table~\ref{table:summary-bregman-div} gives the updates for Frobenius norm (Fro.), KL-divergence (KL), and  Itakura-Saito divergence (IS).
Note that Beta divergence and many others are also easily adapted for our higher-order network modeling framework.

\subsection{Model Selection}
\label{sec:model-learning}
\noindent
In this section, we introduce our approach for learning the appropriate model given an arbitrary graph.
The approach is leverages the Minimum Description Length (MDL)~\cite{grunwald2007minimum,rissanen1978modeling} principle for automatically selecting the ``best'' higher-order network model.
The MDL principle is a practical formalization of Kolmogorov complexity~\cite{li2009introduction}.
More formally, the approach finds the model $\mathcal{M}_{\star} = (\V_{r}, \U_{r})$  that leads to the best compression by solving: 
\begin{align}\label{eq:MDL-opt}
M_{\star} \, = \, \argmin\limits_{M \in \mathcal{M}} \; 
\mathcal{L}_{}(M) \, + \, \mathcal{L}_{}(\mX \,|\, M)
\end{align}
\noindent 
where $\mathcal{M}$ is the model space, $M_{\star}$ is the model given by the solving the above minimization problem, and $\mathcal{L}_{}(M)$ as the number of bits required to encode $M$ using code $\Omega$, which we refer to as the description length of $M$ with respect to $\Omega$. 
Recall that MDL requires a lossless encoding.
Therefore, to reconstruct $\mX$ \emph{exactly} from $M=(\U_r,\V_r)$ we must explicitly encode the error $\mE$ such that 
\[ 
\mX = \U_r\V_r^{T}+\mE
\]
Hence, the total compressed size of $M=(\U_r,\V_r)$ with $M \in \mathcal{M}$ is simply 
$\mathcal{L}(\X, M) = \mathcal{L}(M) + \mathcal{L}(\mE)$.
Given an arbitrary model $M = (\U_r,\V_r) \in \mathcal{M}$, the description length is decomposed into:
\begin{itemize}
\item Bits required to describe the model 
\item Cost of describing the approximation errors 
$\mX - \mX_{r} = \mU_{r}\mV_{r}^{T}$
where $\mX_{r}$ is the rank-r approximation of $\mX$, 
\begin{align}
\mU_{r} & = \big[\, \vu_1 \;\, \vu_2 \;\, \cdots \;\, \vu_r \, \big] \in \RR^{\m \times r}, \quad \text{ and } \\
\mV_{r} &= \big[\, \vv_1 \;\, \vv_2 \;\, \cdots \;\, \vv_r \, \big] \in \RR^{f \times r}
\end{align}
\end{itemize}
The model $M_{\star}$ is the model $M \in \mathcal{M}$ that minimizes the total description length: 
the model description cost $X$ and 
the cost of correcting the errors of our model. 
Let $\abs{\mU}$ and $\abs{\mV}$ denote the number of nonzeros in $\mU$ and $\mV$, respectively.
Thus, the model description cost of $M$ is: 
$\kappa r( \abs{\mU} + \abs{\mV})$ where $\kappa$ is the bits per value.
Similarly, if $\mU$ and $\mV$ are dense, then the model description cost is simply $\kappa r (\m + f)$ where $\m$ and $f$ are the number of edges and features, respectively. 
Assuming errors are non-uniformly distributed, one possibility is to use KL divergence (see Table~\ref{table:summary-bregman-div}) for the error description cost\footnote{The representation cost of correcting approximation errors}.
The cost of correcting a single element in the approximation is $\D{x}{x^{\prime}} = x \log \frac{x}{x^{\prime}} - x + x^{\prime}$ (assuming KL-divergence), and thus, the total reconstruction cost is: 
\begin{align} \label{eq:error-reconstruction-cost-KL}
\D{\mX}{\mX^{\prime}} = \sum\limits_{ij} X_{ij} \log \frac{X_{ij}}{X^{\prime}_{ij}} - X_{ij} + X^{\prime}_{ij}
\end{align}
\noindent
where $\mX^{\prime} = \mU\mV^{T} \in \RR^{\m \times f}$. 
Other possibilities are given in Table~\ref{table:summary-bregman-div}.
The above assumes a particular representation scheme for encoding the models and data.
Recall that the optimal code assigns $\log_2 p_i$ bits to encode a message~\cite{shannon1948mathematical}.
Lloyd-Max quantization~\cite{max1960quantizing,lloyd1982least} with Huffman codes~\cite{huffman1952method,van1976construction} are used to compress the model and data~\cite{oliver1948philosophy,bennett1948spectra}.
Notice that we require only the length of the description using the above encoding scheme, and thus we do not need to materialize the codes themselves.
This leads to the improved model description cost:
$\bar{\kappa} r (\abs{\mU} + \abs{\mV})$ where $\bar{\kappa}$ is the mean bits required to encode each value\footnote{Note $\log_2(\m)$ quantization bins are used}. 
In general, our higher-order network modeling framework can easily leverage other model selection techniques such as AIC~\cite{akaike1974new} and BIC~\cite{schwarz1978estimating}.

\section{Dynamic Edge Role Model} \label{sec:dynamic-edge-role-model} 
\noindent
This section introduces the \emph{dynamic edge role mixed-membership model} (DERM) and proposes a computational framework for computing edge roles in dynamic networks.

\subsection{Dynamic Graph Model \& Representation} \label{sec:dynamic-graph-model-and-representation}
Given a graph stream $G = (V,E)$ where $E=\{e_1, \ldots, e_{\m}\}$ is an ordered set of edges in the graph stream such that $\tau(e_1) \leq \tau(e_2) \leq \cdots \leq \tau(e_{m})$. 
Note that $\tau(e_i)$ is the edge time for $e_i \in E$ (which may be the edge activation time, arrival time, among other possibilities).
Intuitively, $E$ is an infinite edge streaming network where edges arrive continuously over time.
From this edge stream, we derive a dynamic network $\mathcal{G}=\{G_t\}_{t=1}^{T}$ where $G_t = (V, E_t)$ represents a snapshot graph at time $t$. 
Note that time $t$ is actually a discrete time interval $[a,b)$ where $a$ and $b$ are the start and end time, respectively.
Therefore, $E_t = \{e_t \in E \; | \; a \leq \tau(e_i) < b\}$ and $E = E_1 \cup E_2 \cup \cdots \cup E_{T}$.

\subsection{Dynamic Edge Role Learning} \label{sec:dynamic-role-definitions}
We start by learning a time series of features automatically. 
Let $G_{1:k} = (V, E_{1:k})$ be the initial dynamic training graph where $E_{1:k} = E_1 \cup \cdots \cup E_k$ and $k$ represents the number of snapshot graphs to use for learning the initial set of (representative) dynamic features.
Given $\{G_t\}_{t=1}^{T}$ and $G_{1:k} = (V, E_{1:k})$, the proposed approach automatically learns a set of features $\mathcal{F} = \{f_1, f_2, \ldots, f_d\}$ where each $f_i \in \mathcal{F}$ represents a learned feature definition from $G_{1:k}$. 
Given the learned role definitions $\mV \in \RR^{r \times d}$ using a subset of past temporal graphs, 
we then estimate the edge role memberships $\{\mU_t\}_{t=1}^{T}$ for each $\{G_t\}_{t=1}^{T}$ (and any future graph snapshots $G_{t+1}, \ldots, G_{t+p}$) where $\mU_t \in \RR^{\m \times r}$ is an edge by role membership matrix. 
The dynamic edge role model is selected using the approach proposed in Section~\ref{sec:model-learning}.

\subsubsection{Time-scale Learning} \label{sec:time-scale-learning}
This section briefly introduces the problem of learning an appropriate time-scale automatically and proposes a few techniques.
The time-scale learning problem can be formulated as an optimization problem where the optimal solution is the one that minimizes the objective function. 
Naturally, the objective function encodes the error from models learned using a particular time-scale $s$ (\eg, 1 minute, 1 hour).
Thus solving the optimization problem leads to identifying models from the time-scale $s$ that lead to the least error.

\subsubsection{Updating Features and Role Definitions} \label{sec:role-drift-and-updating}
To prevent the features and role definitions from becoming stale and meaningless over time (due to temporal/concept drift as the network and its attributes/properties evolve), we use the following approach:
the loss (or another measure) is computed and tracked over time, and when it becomes too large (from either the features or roles), 
we then re-compute the feature definitions $\mathcal{F}$ and role definitions.
Both the features and roles definitions can be learned in the background as well, and can obviously be computed in parallel.
The edge role framework is also flexible for other types of approaches, and thus, not limited to the simple approach above (which is a key advantage of this work).

\begin{figure}[t!]
\centering
\hspace{-6mm}
\includegraphics[width=0.7\linewidth, bb=0 160 600 640]{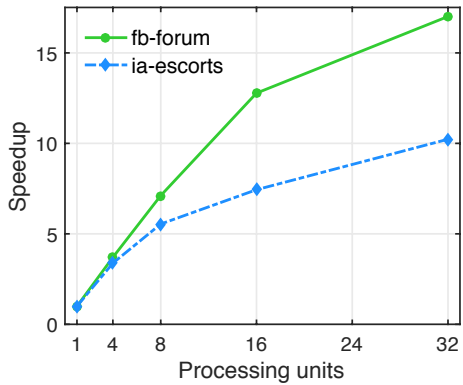}
\caption{
Higher-order role discovery shows strong scaling as we increase the number of processing units.
}
\label{fig:parallel-speedup-feat-edge-roles}
\end{figure}

\section{Experiments} \label{sec-exp}
This section investigates the scalability and effectiveness of the higher-order latent space modeling framework.

\subsubsection{Scalability}
\label{sec:parallel-scaling}
We investigate the scalability of the parallel framework for modeling higher-order latent edge roles. 
To evaluate the effectiveness of the parallel modeling framework, 
we measure the speedup defined as simply $S_p = T_1 / T_p$ where $T_1$ is the execution time of the sequential algorithm, 
and $T_p$ is the execution time of the parallel algorithm with $p$ processing units. 
Overall, the methods show strong scaling (See Figure~\ref{fig:parallel-speedup-feat-edge-roles}). 
Similar results were observed for other networks. 
As an aside, the experiments in Figure~\ref{fig:parallel-speedup-feat-edge-roles} used a 4-processor Intel Xeon E5-4627 v2 3.3GHz CPU.

\begin{figure}[b!]
\centering
\includegraphics[width=1.0\linewidth]{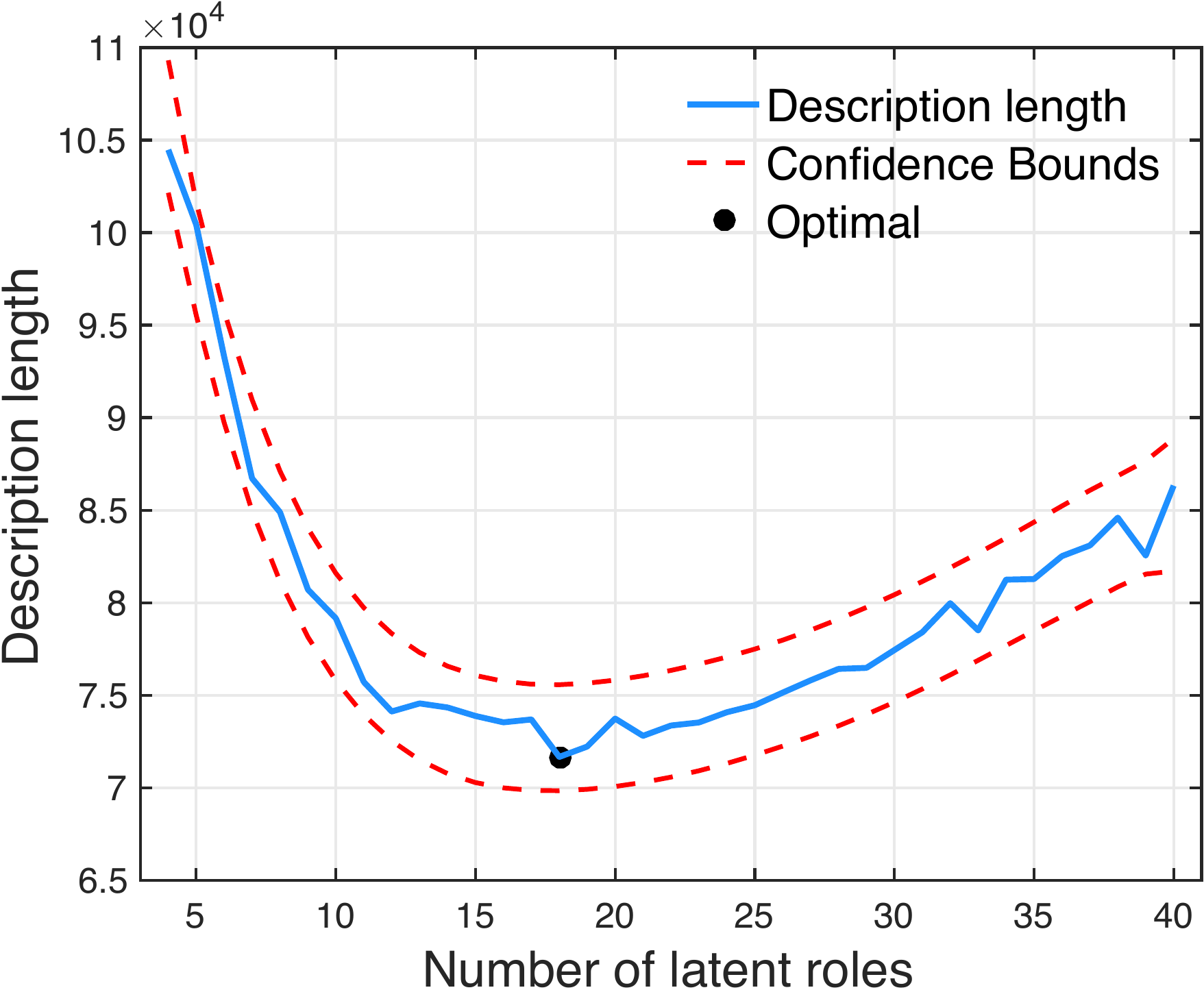}
\caption{
In the example shown, the valley identifies the correct number of latent roles.}
\label{fig:MDL-edge-roles}
\end{figure}

\subsubsection{Higher-order Model Selection}
\begin{figure}[t!]
\centering
\includegraphics[width=0.8\linewidth, bb=0 160 600 620]{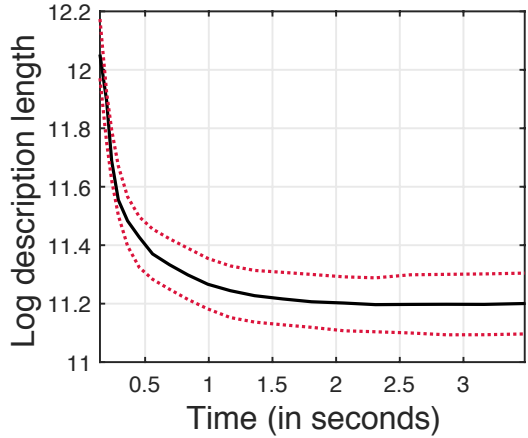}
\caption{The running time of our approach. The x-axis is time in seconds and the y-axis is the log description cost. 
The curve is the average over 50 experiments and the dotted lines represent three standard deviations. 
The result reported above is from a laptop with a single core.
}
\label{fig:runtime-vs-desc-cost}
\end{figure}

\begin{figure*}[t!]
        \centering
        \subfigure[Learning time]{
        		    \includegraphics[width=0.30\linewidth]{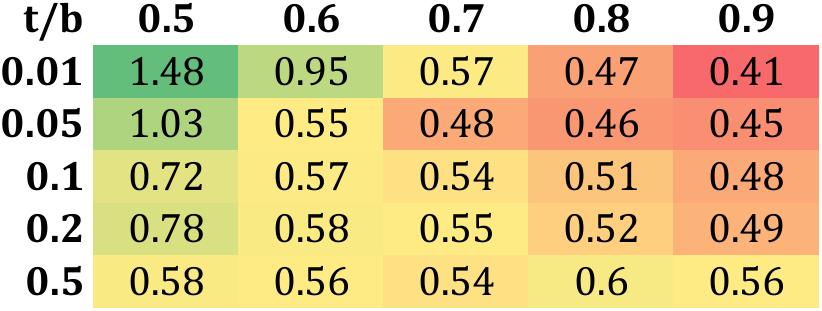}
                \label{fig:learning-time}}
        \subfigure[Number of features discovered]{
        		    \includegraphics[width=0.30\linewidth]{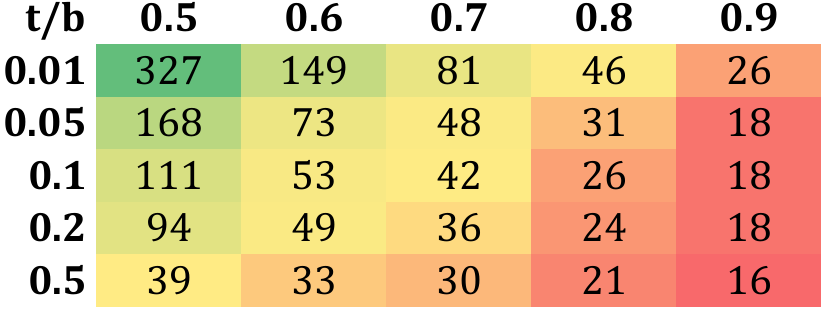}
                \label{fig:number-of-features}}
        \subfigure[Sparsity of features]{
        		    \includegraphics[width=0.30\linewidth]{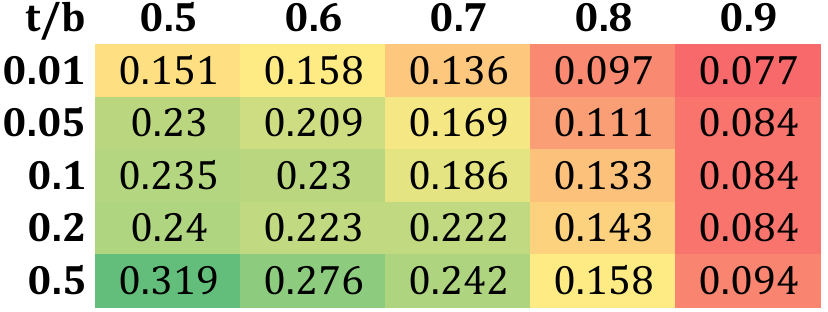}
                \label{fig:sparsity-of-features}}
        \caption{
        Impact on the learning time, number of features, and their sparsity, as the tolerance ($\varepsilon$) and bin size ($\alpha$) varies.
        }
        \label{fig:varying-tol-and-binsize}
\end{figure*}

MDL is used to automatically learn the appropriate edge role model.
In Figure~\ref{fig:MDL-edge-roles}, description length (in bits) is minimized when $r=18$.
Intuitively, too many roles increases the model description cost, whereas too few roles increases the cost of describing errors. 
In addition, Figure~\ref{fig:runtime-vs-desc-cost} shows the runtime of our approach.
Furthermore, Figure~\ref{fig:varying-tol-and-binsize} demonstrates the impact on the learning time, number of novel features discovered, and their sparsity, as the tolerance ($\varepsilon$) and bin size ($\alpha$) varies.

\subsubsection{Modeling Dynamic Networks}
\label{sec:exp-modeling-dynamic-networks}
In this section, we investigate the Enron email communication networks using the \emph{Dynamic Edge Role Mixed-membership Model} (DERM). 
The Enron email data consists of 151 Enron employees whom have sent 50.5k emails to other Enron employees. 
We processed all email communications spanning over 3 years of email communications, and discarded the textual content of the email, and only use the edges representing a directed email communication (from one employee to another). 
The email communications are from 05/11/1999 to 06/21/2002. 

\begin{figure}[t!]
\centering
\subfigure[Evolving \emph{edge role} mixed-memberships]{\includegraphics[width=0.940\linewidth]{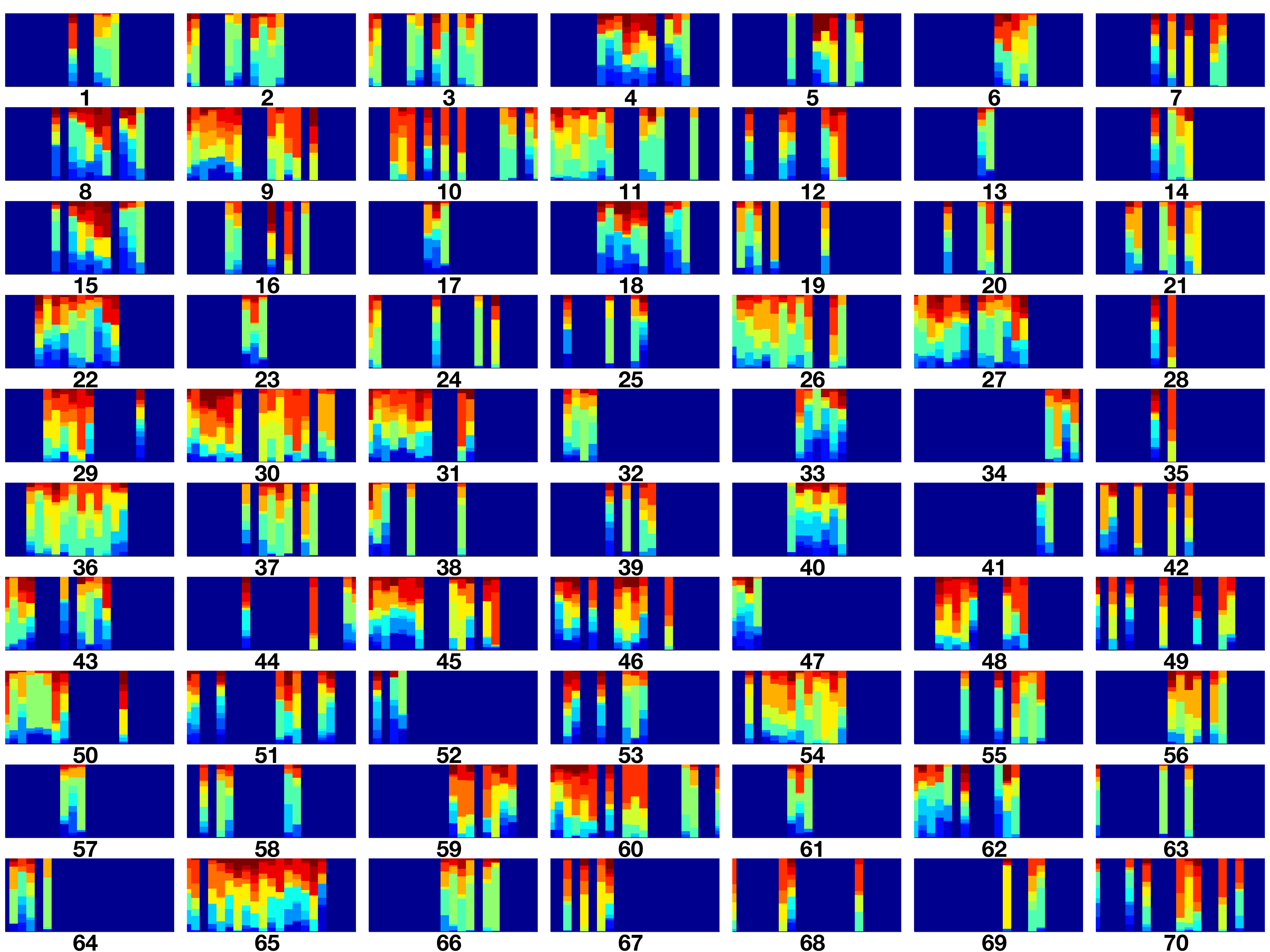} \label{subfig:dynamic-edge-roles-enron}}
\subfigure[Evolving \emph{node role} mixed-membership]{\includegraphics[width=0.940\linewidth]{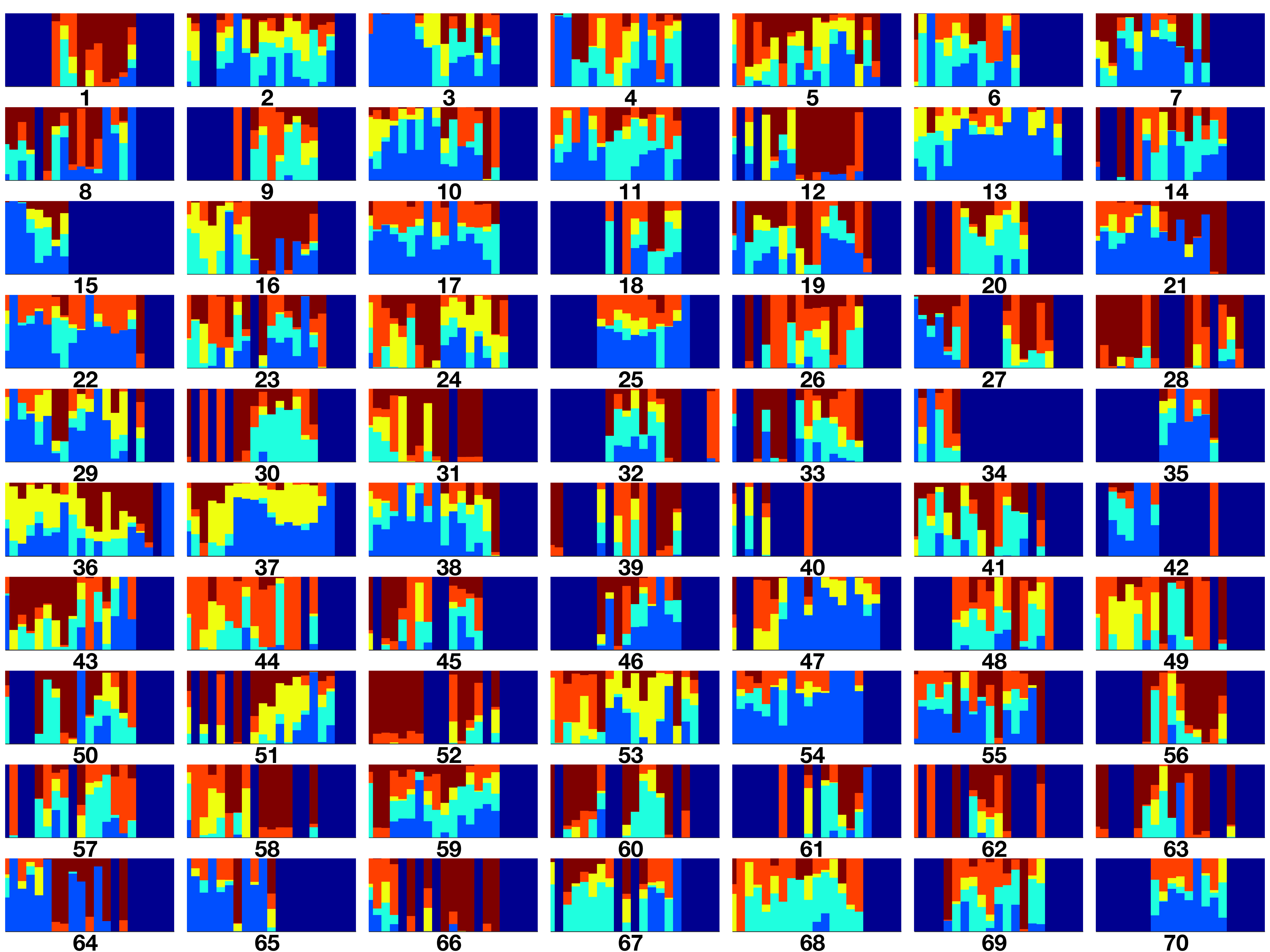} \label{subfig:dynamic-node-roles-enron}}
\vspace{-1mm}
\caption{
Temporal changes in the edge and node mixed-membership vectors (from the Enron email communication network). 
The horizontal axes of each subplot is time, whereas the vertical axes represent the components of each mixed-membership vector. 
Roles are represented by different colors. 
}
\label{fig:dynamic-edge-roles}
\end{figure}

For learning edge roles (and a set of representative edge features), we leverage the first year of emails. 
Note that other work such as dMMSB~\cite{fu2009dynamic} use email communications from 2001 only, which corresponds to the time period that the Enron scandal was revealed (October 2001).
We instead study a much more difficult problem. In particular, given only past data, can we actually uncover and detect the key events leading up to the downfall of Enron?
A dynamic network $\{G_t\}^{T}_{t=1}$ is constructed from the remaining email communications (approximately 2 years) where each snapshot graph $G_t$, $t=1,\ldots,T$ represents a month of communications. 
Interestingly, we learn a \emph{dynamic node role mixed-membership model} with 5 latent roles, which is exactly the number of \emph{latent node roles} learned by dMMSB~\cite{fu2009dynamic}. 
However, we learn a dynamic \emph{edge} role mixed-membership model with 18 roles.
Evolving edge and node mixed-memberships from the Enron email communication network are shown in Figure~\ref{fig:dynamic-edge-roles}.
The set of edges and nodes visualized in Figure~\ref{fig:dynamic-edge-roles} are selected using the difference entropy rank (See Eq.\eqref{eq:diff-entropy-rank} below) and correspond to the edges and nodes with largest difference entropy rank $\vd$. 
The first role in Figure~\ref{fig:dynamic-edge-roles} represents inactivity (dark blue).

For identifying anomalies, we use the difference entropy rank defined as:
\begin{align} \label{eq:diff-entropy-rank}
\vd = \max_{t\in T} H(\vu_{t}) - \min_{t\in T} H(\vu_{t})
\end{align}
\noindent
where $H(\vu_{t}) = - \vu_t \cdot \log(\vu_t)$ and $\vu_t$ is the $r$-dimensional mixed-membership vector for an edge (or node) at time $t$.
Using the difference entropy rank, we are able to reveal important communications between key players involved in the Enron Scandal, such as Kenneth Lay, Jeffrey Skilling, and Louise Kitchen. 
Notice that when node roles are used for identifying dynamic anomalies in the graph, we are only provided with potentially malicious employees, whereas using edge roles naturally allow us to not only detect the key malicious individuals involved, but also the important relationships between them, which can be used for further analysis, among other possibilities.

\subsubsection{Exploratory Analysis}
\begin{figure}[b!]
\centering
\includegraphics[width=1.0\linewidth]{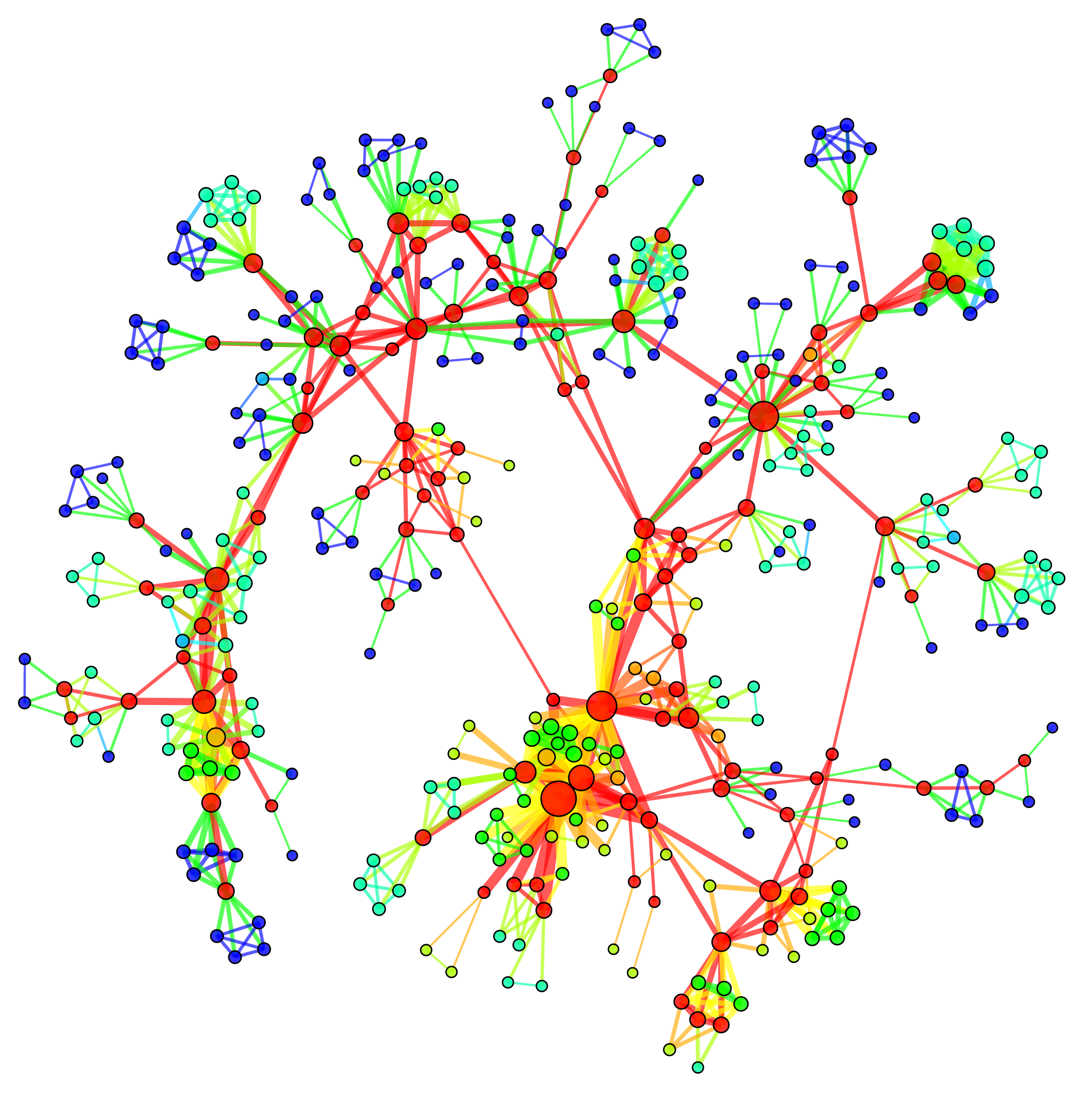}
\caption{Edge and node roles for ca-netscience.
Link color represents the edge role and node color indicates the corresponding node role. 
}
\label{fig:node-and-edge-roles-ca-netscience}
\end{figure}

Figure~\ref{fig:node-and-edge-roles-ca-netscience} visualizes the node and edge roles learned for ca-netscience.
While our higher-order latent space model learns a stochastic $r$-dimensional vector for each edge (and/or node) representing the individual role memberships, Figure~\ref{fig:node-and-edge-roles-ca-netscience} assigns a single role to each link and node for simplicity.
In particular, given an edge $e_i \in E$ (or node) and its mixed-membership row vector $\vu_i$, we assign $e_i$ the role with maximum likelihood $k_{\star} \leftarrow \argmax_{k} u_{ik}$.
The higher-order edge and node roles from Figure~\ref{fig:node-and-edge-roles-ca-netscience} are clearly meaningful. 
For instance, the red edge role represents a type of bridge relationship as shown in Figure~\ref{fig:node-and-edge-roles-ca-netscience}.

\subsubsection{Sparse Graph Feature Learning} 
\label{sec:sparse-graph-feature-learning}
Recall that the proposed feature learning approach attempts to learn ``sparse graph features'' to improve learning and efficiency, especially in terms of space-efficiency. 
This section investigates the effectiveness of our sparse graph feature learning approach.
Results are presented in Table~\ref{table:sparse-graph-features}. 
In all cases, our approach learns a highly compressed representation of the graph, requiring only a fraction of the space of current (node) approaches.
Moreover, the density of edge and node feature representations learned by our approach is between $[0.164, 0.318]$ and $[0.162, 0.334]$ for nodes (See $\rho(\mX)$ and $\rho(\mZ)$ in Table~\ref{table:sparse-graph-features}) and up to $6x$ more space-efficient than other approaches.
While existing feature learning approaches for graphs are unable to learn higher-order graph features (and thus impractical for higher-order network analysis and modeling), 
they also have another fundamental disadvantage: they return dense features.
Learning space-efficient features is critical especially for large networks. 
For instance, notice that on extremely large networks, storing even a small number of edge (or node) features quickly becomes impractical. 
Despite the importance of learning sparse graph features, existing work has ignored this problem as most approaches stem from Statistical Relational Learning (SRL)~\cite{introSRL07} and have been designed for extremely small graphs.
Moreover, nearly all existing methods focus on node features~\cite{davis2007changeSAYU-VISTA,kok2007statistical,landwehr2006kfoil-new,landwehr2005nfoil}, whereas we focus on both and primarily on learning novel and important edge feature representations from large massive networks.

\begin{table}[h!]
\caption{
Higher-order sparse graph feature learning for latent node and edge network modeling.
Recall that $f$ is the number of features, 
$L$ is the number of layers, and 
$\rho(\mX)$ is the sparsity of the feature matrix. 
Edge values are bold.
}
\vspace{1mm}
\label{table:sparse-graph-features}
\setlength{\tabcolsep}{4pt} 
\centering 
\small
\def\arraystretch{1.2}
\begin{tabularx}{1.0\linewidth}{@{}
r
X
llX
llX
llH
}
\toprule
\textbf{graph}  &	 
$f$  &&& $L$ &&&	 $\rho(\mX)$ &$\rho(\mZ)$ &
\\
\midrule
\def\arraystretch{2.5}

\textrm{socfb-MIT} &	
 \textbf{2080} & (912) &&	 \textbf{8} & (9) &&	 \textbf{0.318} & (0.334) 
\\

\textrm{yahoo-msg} &	 
\textbf{1488} & (405) &&	 \textbf{7} & (7) &&	 \textbf{0.164} & (0.181) 
\\

\textrm{enron} &	 
\textbf{843} & (109) &&	 \textbf{5} & (4) &&	 \textbf{0.312} & (0.320) 
\\

\textrm{Facebook} &	 
\textbf{1033} & (136) &&	\textbf{7} & (5) &&	\textbf{0.187} & (0.162) 
\\

\textrm{bio-DD21} &	 
\textbf{379} & (723) &&	 \textbf{6} & (6) &&	 \textbf{0.215} & (0.260) 
\\
\bottomrule
\end{tabularx}
\end{table}

\subsubsection{Computational Complexity}
\label{sec-complexity}
Recall that $\m$ is the number of edges, $\n$ is the number of nodes, $f$ is the number of features, and $r$ is the number of latent roles.
The total time complexity of the \emph{higher-order latent space model} is: 
$\mathcal{O}\big(f (\m + nr)\big)$.
\noindent
Thus, the runtime is linear in the number of edges.
The time complexity is decomposed into the following main parts:
Feature learning takes $\mathcal{O}(f(m+nf))$.
Model learning takes $\mathcal{O}(\m\r\f)$ in the worst case (which arises when $\mU$ and $\mV$ are completely dense). 
The quantization and Huffman coding terms are very small and therefore ignored.
Latent role learning using scalar element-wise coordinate descent has worst case complexity of $\mathcal{O}(\m f r)$ per iteration which arises when $\mX$ is completely dense.
However, assuming $\mX$ is sparse, then it takes $\mathcal{O}(\abs{\mX} r)$ per iteration where $\abs{\mX}$ is the number nonzeros in $\mX \in \RR^{\m \times f}$.
In addition, we compute the initial set of graphlet-based features using the efficient parallel algorithm in~\cite{ahmed2016kais}. Note that this algorithm computes the counts of a few graphlets and directly obtain the others in constant time. This takes $\mathcal{O}( \dmax \big(|S_u|+|S_v|+|T_{\e}|\big))$ for any given edge $e_i = (v,u)$, where $\dmax$ is the maximum degree for any vertex, $S_v$, $S_u$ are the sets of wedge nodes and $T_{\e}$ is the set of triangles incident to edge $e_i$.

\section{Conclusion} \label{sec-conc}
\noindent
This work introduced the notion of edge roles and proposed a higher-order latent space network model for edge role discovery. 
To the best of our knowledge, this work is the first to explore using higher-order graphlet-based features for role discovery.
Moreover, these features are counts of various induced subgraphs of arbitrary size and were used directly for role discovery as well as given as input into a graph representation learning approach to learn more discriminative features based on these initial features.
Furthermore, feature-based edge roles also have many important and key properties and can be used for graph similarity, node and edge similarity queries, visualization, anomaly detection, classification, link prediction, among many other tasks.
Our edge role discovery framework also naturally supports large-scale attributed networks.

\bibliographystyle{abbrv}
\bibliography{paper}

\end{document}